\documentclass[preprints,article,accept,pdftex,moreauthors]{Definitions/mdpi} 
\firstpage{1} 
\pubvolume{1}
\issuenum{1}
\articlenumber{0}
\pubyear{2022}
\copyrightyear{2022}
\datereceived{} 
\dateaccepted{} 
\datepublished{} 
\hreflink{https://doi.org/} 
\pdfoutput=1

\usepackage{amsmath,amssymb,amsfonts}
\usepackage{algorithmic}
\usepackage{graphicx}
\usepackage{textcomp}
\usepackage{multirow}
\usepackage{boldline}
\usepackage{stackengine}
\usepackage{siunitx}
\usepackage{enumitem}
\usepackage{physics}
\usepackage{booktabs}
\usepackage{makecell}
\newcommand{\rf}[1]{\cellcolor{black!20}{#1}}
\renewcommand{\bf}[1]{\cellcolor{black!20}\textbf{#1}}
\usepackage{romannum}
\usepackage[ruled,vlined]{algorithm2e}
\usepackage{subcaption}

\usepackage[acronym]{glossaries}

\newacronym{ES-RNN}{ES-RNN}{Exponential Smoothing - Recurrent Neural Network}
\newacronym{FFORMA}{FFORMA}{Feature-based FORecast Model Averaging}
\newacronym{N-BEATS}{N-BEATS}{Neural Basis Expansion Analysis}
\newacronym{RF}{RF}{Random Forests}
\newacronym{XGBoost}{XGBoost}{eXtreme Gradient Boosting Machine}
\newacronym{MLP}{MLP}{Multilayer Perceptron}
\newacronym{ANN}{ANN}{Artificial Neural Network}
\newacronym{RNN}{RNN}{Recurrent Neural Network}
\newacronym{LSTM}{LSTM}{Long Short Term Memory}
\newacronym{ARIMA}{ARIMA}{AutoRegressive Integrated Moving Average}
\newacronym{ML}{ML}{Machine Learning}
\newacronym{OWA}{OWA}{Overall Weighted Average}
\newacronym{sMAPE}{sMAPE}{Symmetric Mean Absolute Percentage Error}
\newacronym{MAPE}{MAPE}{Mean Absolute Percentage Error}
\newacronym{MASE}{MASE}{Mean Absolute Scaled Error}
\newacronym{ES}{ES}{Exponential Smoothing}
\newacronym{FFORMS-R}{FFORMS-R}{Feature-based FORecast Model Selection using Random Forest}
\newacronym{FFORMS-G}{FFORMS-G}{Feature-based FORecast Model Selection using Gradient Boosting}
\newacronym{FFORMA-N}{FFORMA-N}{Feature-based FORecast Model Averaging using Neural Networks}
\newacronym{NN-STACK}{NN-STACK}{Neural Networks Stacking Regression}
\newacronym{ReLU}{ReLU}{Rectified Linear Unit}
\newacronym{AVG}{AVG}{Simple Model Averaging}

\DeclareMathOperator*{\argmin}{arg\,min}

\Title{Evaluating State of the Art, Forecasting Ensembles- and Meta-learning Strategies for Model Fusion}

\TitleCitation{Evaluating State of the Art, Forecasting Ensembles- and Meta-learning Strategies for Model Fusion}

\Author{Pieter Cawood $^{1}$\orcidA{}, Terence van Zyl $^{2}$\orcidB{}}

\AuthorNames{Firstname Lastname, Firstname Lastname and Firstname Lastname}

\AuthorCitation{Cawood, P.; van Zyl, T.}

\address{%
$^{1}$ \quad Institute for Intelligent Systems, University of Johannesburg, Johannesburg, South Africa; pieter.cawood@gmail.com\\
$^{2}$ \quad Institute for Intelligent Systems, University of Johannesburg, Johannesburg, South Africa; tvanzyl@uj.ac.za}

\abstract{Techniques of hybridisation and ensemble learning are popular model fusion techniques for improving the predictive power of forecasting methods. With limited research that instigates combining these two promising approaches, this paper focuses on the utility of the \gls{ES-RNN} in the pool of base learners for different ensembles. We compare against some state of the art ensembling techniques and arithmetic model averaging as a benchmark. We experiment with the M4 forecasting dataset of $100,000$ time-series, and the results show that the \acrlong{FFORMA}, on average, is the best technique for late data fusion with the \gls{ES-RNN}. However, considering the M4's Daily subset of data, stacking was the only successful ensemble at dealing with the case where all base learner performances are similar. Our experimental results indicate that we attain state of the art forecasting results compared to \acrlong{N-BEATS} as a benchmark. We conclude that model averaging is a more robust ensembling technique than model selection and stacking strategies. Further, the results show that gradient boosting is superior for implementing ensemble learning strategies.}


\begin{document}




\section{Introduction}
\glsresetall
Forecasting is the procedure of creating predictions based on past and current data. Subsequently, these predictions can be compared against what happened. For example, one might estimate future infections and then compare them against the actual outcomes. Forecasting might refer to specific formal statistical methods employing time-series, cross-sectional or longitudinal data, less traditional judgmental methods or the process of prediction and resolution itself~\cite{cawood2021feature}.

Although challenging, forecasting time-series is an essential task to which substantial research effort has been applied. Makridakis \emph{et al.} \cite{makridakis2020forecasting} emphasise two facts about the field: first, no one has prophetic powers to predict the future accurately, and second, all predictions are subject to uncertainty, especially within social contexts.

Forecasting has applications in many fields where estimates of future conditions are helpful. Depending on the area, accuracy varies significantly. If the factors that relate to the forecast are known and well understood and a significant amount of data can be used, the final value will likely be close to the estimates. If this is not the case or if the actual outcome is affected by the forecast, the dependability of the predictions can be significantly lower. Some forecasting applications include Climate change, preventative maintenance, anomalies, stock returns, epidemics, economic trends, and the development of conflict situations~\cite{atherfold2020method,mathonsi2022multivariate,freeborough2022investigating,timilehin2021surrogate}.

\subsection{Model Fusion and Meta-learning}
In this section we refine the concepts of meta-learning and model fusion and present a brief framework to better contextualise the later literature review as it relates to the presented forecasting techniques. First, we give a definition of model fusion as it relates to forecasting and how it differs from traditional data fusion. Next we describe what meta-learning is within the context of model fusion for time-series forecasting.

Data fusion is the multi-modal, multi-resolution, multi-temporal process of integrating data sources to produce more consistent, accurate, and useful information than the sources individually. Traditional multi-modal data fusion approaches are grouped into classes, based on the processing level at which the fusion occurs. Early fusion or data-level fusion, late fusion or decision level fusion and intermediate fusion which combines late and early approaches~\cite{michelsanti2021overview}.

Analogously, we define \textbf{model fusion} as integrating base learners to produce a lower biased and variance; and more robust, consistent, and accurate model than the learners individually. These base learners can either be \textbf{homogeneous} from the same hypothesis class (e.g. decisions trees in a random forest) or be \textbf{heterogeneous} from different hypothesis classes (e.g. neural network with a support vector machine for the classification). Drawing further from multi-modal data fusion which sub-categorises based on the whether or not the fusion process is early or late we present definitions for early model fusion, late model fusion and incremental model fusion. We define these terms as follows:
\begin{description}
    \item[\textbf{Early model fusion}] integrates the base learners before training. The combined model is then trained as a single fused model.
    \item[\textbf{Late model fusion}] first trains the base learners individually. The now pre-trained base learners are then integrated without further modification. 
    \item[\textbf{Incremental model fusion}] performs model integration while training the base learners incrementally. Each combined base learner's parameters remains fixed once trained.
\end{description}

An important aspect of model fusion is that of meta-learning. Meta-learning is typically classified into metric-based, model-based, and optimisation-based meta-learning. Of interest in our discussion is model-based meta-learning. Model-based meta-learning uses metadata/meta-features about the problem together with a meta-model to improve overall performance. For instance, stacked generalisation works by using linear-regression to weight heterogeneous base learners~\cite{arinze1994selecting}. We are now in a position to define the model integration process:
\begin{description}
    \item[\textbf{Meta-model fusion}] uses model-based meta-learning to perform the model integration process. 
    \item[\textbf{Aggregation fusion}] or just \textbf{aggregation} uses a simple aggregation scheme, like weighted averaging, to perform the model integration.
\end{description}

\begin{figure}
    \centering
    \includegraphics[width=\columnwidth]{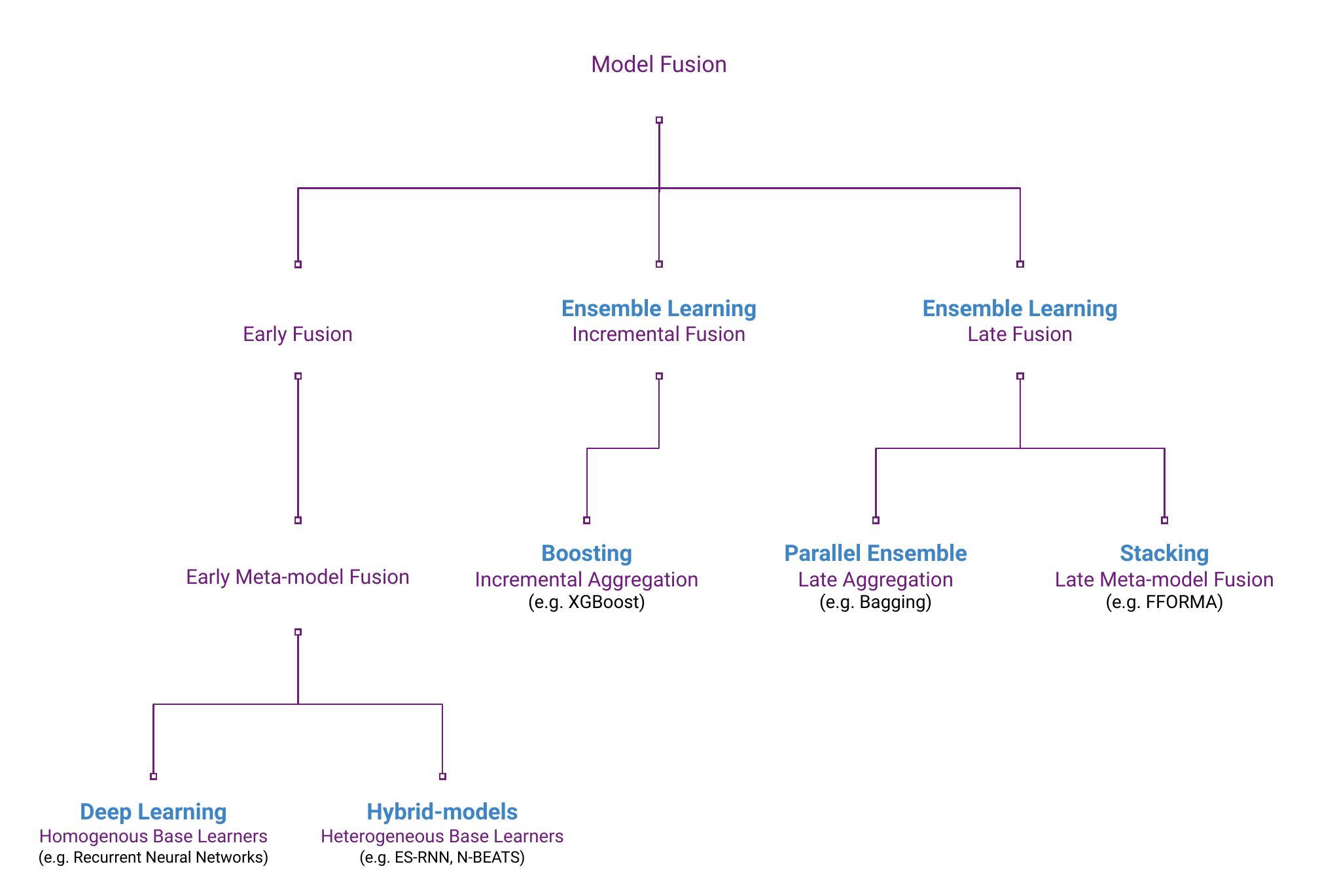}
    \caption{An incomplete taxonomy of model fusion. Showing Ensemble Learning, Boosting, Parallel Ensembles, Stacking, Deep Learning, and Hybrid-models~\cite{ribeiro2020ensemble,zhang2003time}.}
    \label{fig:model_fusion}
\end{figure}

Combing the above discussion with the commonly used appellative for the classes of techniques allows us to arrive at the following \textit{incomplete} taxonomy shown in Figure~\ref{fig:model_fusion}. In the following section we review relevant literature for forecasting model-fusion in the context of the presented taxonomy.

\subsection{Related Literature}
Despite the diversity of the forecasting models, the no free lunch theorem holds that no single algorithm universally outperforms any other for all problems~\cite{wolpert1997no}. As a result, an essential question is raised about choosing the best algorithm for a specific time-series. These many techniques are underpinned by different principles to perform forecasting. However, numerous forecasting studies have shown that a fusion of models generally outperforms the individuals~\cite{reich2019collaborative, mcgowan2019collaborative, johansson2019open}. 

Arinze~\cite{arinze1994selecting} first introduced meta-learning as stacking technique to select one forecasting model among six others based on the learnt performance for six time-series meta-features (e.g., autocorrelation, trend, coefficients). Raftery \emph{et al.} \cite{raftery1997bayesian} point out that single model selection is open to issues of model uncertainty and proposed using Bayesian model averaging for stacking instead. Bayesian model averaging makes weighted average predictions as the posterior probability of each model being correct given the same dataset. Stacking can also learn the weights from the predictions of heterogeneous base learners using cross-validation instead of using posterior probabilities~\cite{clarke2003comparing}. Clarke~\cite{clarke2003comparing} reported cross-validation based stacking as a more robust method than Bayesian model averaging for computations that involve sensitive changes over their variables. 

Stacking methods' performance might also be improved by utilising additional time-series statistics as meta-features in the learning process. Cawood and van Zyl \cite{cawood2021feature}, for example, proposed a feature-weighted stacking approach that improves the regression from base learner predictions by supplementing them with time-series statistics extracted from the input series. Recent work has established many valuable statistics for meta-learning, including, but not limited to, the linearity, curvature, autocorrelation, stability, and entropy~\cite{lorena2018data, barak2019time, montero2020fforma}.

Ribeiro and Coelho \cite{ribeiro2020ensemble} studied contemporary ensemble methods in forecasting time-series within agribusiness. In their work, they compare approaches of bagging with \acrlong{RF}, boosting with gradient boosting machines, \gls{XGBoost} and a stacked generalisation of machine learning models. Ribeiro and Coelhoa's findings suggest that gradient boosting methods generally produce the lowest prediction errors. The top two submissions of the M4 forecasting competition both implement ensemble learning to produce state of the art results. 

The runner-up in the M4 Competition, \gls{FFORMA}, implements stacking using extreme gradient boosting to learn the weightings for the base learners based on several meta-features extracted from the input time-series~\cite{montero2020fforma}. The weightings learned by the trained meta-model are then used to combine the models' predictions as a feature-weighted average.

Hybrid approaches combine linear and nonlinear models since time-series are rarely pure linear or nonlinear in reality \cite{zhang2003time}. Hybrid forecasting models were first proposed by Zhang~\cite{zhang2003time}, who showed that the hybridisation of the \gls{ARIMA} and \gls{MLP} models yields improved accuracy since it takes advantage of combining the strength of both linear- and nonlinear models. The authors note that the main advantage of neural networks is their ability to do nonlinear modelling. In their implementation, the \gls{MLP} learns from the residuals of the linear \gls{ARIMA} method to make the final integrated predictions. The \gls{ES-RNN} is a hybrid forecasting model and winning submission of the M4 Competition that merges a modified Holt-Winters and dilated \gls{LSTM} stacks \cite{ESRNN}. 

Hybrid methods have gained traction in recent time-series forecasting research \cite{liu2014hybrid, wang2015study, qin2019hybrid}. The most common approach is to combine deep learning with a linear method, similar to the initial work by Zhang. Numerous hybrids implement \glspl{ANN} or combine them with traditional models~\cite{mathonsi2020prediction, laher2021deep, mathonsi2022statistics}. Zhang \emph{et al.} \cite{zhang1998forecasting} identify this attraction to \glspl{ANN} as their ability to self-adapt and model underlying relationships within the data, even when the relationships are unknown. However, some have proposed hybridising fuzzy logic techniques with Machine Learning models as well~\cite{aksoy2014demand, deng2015novel}. Further, others have also reported success in integrating Machine Learning models with genetic algorithms like particle swarm, and ant colony optimisation \cite{rahmani2013hybrid, kumar2019novel}. 

Hybrid machine Learning techniques are currently the preferred family for research in forecasting time-series, owing to their predictive accuracy~\cite{shinde2020forecasting}. As it stands, \gls{N-BEATS} achieves state of the art forecasting accuracy by 3\% over the \gls{ES-RNN}~\cite{oreshkin2019n}. \gls{N-BEATS} is a hybrid forecasting method that integrates a fully connected neural network with traditional time-series decomposition. \gls{N-BEATS} proposes a novel doubly residual topology that stacks blocks of different architectures to model the various time-series components. The \gls{N-BEATS} stacking topology resembles the DenseNet architecture proposed by Huang \emph{et al.} \cite{huang2017densely}; however, it has two residual branches that run over both foreword and backward prediction branches of each layer. The neural network thus produces expansion coefficients that are projected to the block basis functions in each layer to produce partial forecasts and backcasts that assist downstream blocks in ignoring components of their inputs that are not useful.

\subsection{Makridakis Competitions (M-Competitions)}
The Makridakis Competitions (or M-Competitions) are undoubtedly the most influential initiative driving continuous research in improving the aforementioned forecasting methods. The M-competitions are a standard benchmark for establishing an innovatory of forecasting methods~\cite{makridakis2020forecasting, makridakis2021m5}. Makridakis \emph{et al.} \cite{makridakis2020forecasting} point out that machine learning models have delivered poor performances for the forecasting of social sciences when considering the M-competitions prior to the recent M5 Competition. 

In the M4 Competition, Makridakis \emph{et al.} \cite{makridakis2020m4} presented a large scale comparative study of $100,000$ time-series and 61 forecasting methods. They compare the accuracy of statistical methods e.g., \gls{ARIMA} and Holt-Winters versus that of Machine Learning methods e.g., \gls{MLP} and \gls{RNN}. Similar to previous studies, they find that utilising hybrid-model fusion of heterogeneous base learners produces the best results~\cite{reich2019collaborative, mcgowan2019collaborative, johansson2019open}. For a comprehensive study of statistical and Machine Learning forecast models, refer to the extensive comparative paper by Makridakis \emph{et al.} \cite{makridakis2020m4}.

More recently, the M5 Competition was the first in the series of M-competitions with hierarchical time-series data \cite{makridakis2021m5}. Subsequently, the M5 Competition led research efforts to confirm the dominance of meta-model fusion in forecasting homogeneous data. The top entries of the M5 Competition, methods that adopt parallel ensemble learning and gradient boosting produce state of the art forecasting results~\cite{makridakis2022m5}. Ensembles are known to outperform their constituent models for many machine learning problems~\cite{rokach2010ensemble}. Likewise, integrating models with different architectures has been reported to deliver superior results \cite{wu2010hybrid, khashei2012new}. 

For the above reasons, this study returns focused to the M4 Competition to investigate if the results from the M5 Competition relating to fusion strategies hold over the forecasting of univariate time-series.

\subsection{Contributions of this Paper} 
A review of time-series forecasting literature highlights the need to determine the improvement one might expect to achieve by implementing late data fusion with the state of the art hybrid time-series forecasting models. Moreover, research is required to compare the performance of different ensembles using the forecasts from the same pool of forecasting models. There is also a requirement in forecasting research that necessitates one to validate the utility of ensemble learning with an advanced hybrid model. Our work is novel in that it is the first study that adopts late meta-model fusion to integrate traditional base learners with a hybrid-model. We contribute to the forecasting literature by:
\begin{enumerate}
    \item presenting a novel taxonomy for organising the current literature around forecasting model fusion;
    \item studying the potential improvement of the predictive power of any state of the art forecasting model;
    \item contrasting the performance of multiple ensembling techniques from different architectures; and
    \item delivering an equitable comparison of techniques by providing validation results of the ensembles over five runs of ten-fold cross-validation.
\end{enumerate}

\section{Materials and Methods}
\glsresetall
In this section we first give a brief overview of the M4 Competition followed by each base learner utilised. The base learners include four statistical ones and the \gls{ES-RNN}, a hybrid deep learning model. We reused the original base learner forecasts from the M4 submissions as the input forecasts for the ensembles. The later part of this section gives an overview of the different ensemble techniques that we implemented for our experiments and how they differ in their complexity and meta-learning strategies.

\subsubsection{The M4 Forecasting Competition}
One instance of the M-Competitions is the M4 forecasting competition. The M4 dataset contains $100,000$ time-series of different frequencies, i.e., yearly, quarterly, monthly, weekly, daily and hourly. The minimum number of observations varies for the different subsets, e.g., 13 for the yearly series and 16 for quarterly. The data is freely available on Github \footnote{https://github.com/Mcompetitions/M4-methods/tree/master/Dataset} and Table~\ref{tab:m4data} provides a summary of the number of series per frequency and domain. Economic, Finance, Demographics, and Industries are among the domains, with data from Tourism, Trade, Labor and Wage, Real Estate, Transportation, Natural Resources, and the Environment also included. 
\begin{table}[!ht]
\centering
\caption{Number of series per data frequency and domain \cite{makridakis2020m4}.}
\begin{tabular}{l|rrrrrrr}
    Data subset 
    & Micro
    & Industry
    & Macro
    & Finance
    & \makecell[b]{Demo-\\graphic}
    & Other
    & Total\\ \bottomrule\toprule
Yearly    & 6,538  & 3,716  & 3,903  & 6,519  & 1,088 & 1,236 & 23,000  \\ 
Quarterly & 6,020  & 4,637  & 5,315  & 5,305  & 1,858 & 865   & 24,000  \\ 
Monthly   & 10,975 & 10,017 & 10,016 & 10,987 & 5,728 & 277   & 48,000  \\ 
Weekly    & 112    & 6      & 41     & 164    & 24    & 12    & 359  \\ 
Daily     & 1,476  & 422    & 127    & 1,559  & 10    & 633   & 4,227  \\ 
Hourly    & 0      & 0      & 0      & 0      & 0     & 414   & 414  \\ 
Total     & 25,121 & 18,798 & 19,402 & 24,534 & 8,708 & 3,437 & 100,000  \\
\bottomrule
\end{tabular}
\label{tab:m4data}
\end{table}

The forecast horizon lengths for each frequency are six steps ahead for yearly data, eight steps ahead for quarterly data, 18 steps ahead for monthly data, 13 steps ahead for weekly data, 14 steps ahead for daily data, and 48 forecasts for hourly data. The M4 Competition comprises one essential considerations relating to \gls{OWA} as a metric for comparison detailed further next.

The \gls{OWA} computes the average of the two most popular accuracy measures to evaluate the performance of the PFs relative to the Naïve model 2. The \gls{OWA}, therefore, combines the two metrics of the \gls{sMAPE} \cite{makridakis1993accuracy} and mean absolute scaled error \gls{MASE} \cite{hyndman2006another}, each calculated as follows:
\begin{equation}
\begin{aligned}
\label{formula:smape}
sMAPE = \frac{1}{h} \sum^{h}_{t=1} \frac{2|Y_t - \hat{Y_t}|}{|Y_t| + |\hat{Y_t}|} \times 100\%,
\end{aligned}
\end{equation}
\begin{equation}
\begin{aligned}
\label{formula:mase}
MASE = \frac{1}{h} \frac{\sum^{n+h}_{t=n+1}|Y_t - \hat{Y_t}|}{\frac{1}{n-m}\sum^{n}_{t=m+1}|Y_t - Y_{t-m}|},
\end{aligned}
\end{equation}

where $Y_t$ is the actual time-series value at time step $t$, $\hat{Y}_t$ is the predicted value of $Y_t$ and $h$ the length of the forecasting horizon. The denominator and scaling factor of the \gls{MASE} formula are the in-sample mean absolute error from one-step-ahead predictions of the Naïve model 2, and $n$ is the number of data points. The term $m$ defines the time interval between each successive observation, i.e., $12$ for time-series that have a monthly frequency, four for those with a quarterly frequency, $24$ for hourly series and one for the other frequencies that are nonseasonal series. The \gls{OWA} error is then computed as the average of the \gls{MASE} and \gls{sMAPE} errors relative to the Naïve model 2 predictions as follows:

\begin{equation}
\begin{aligned}
\label{formula:owa}
OWA = \frac{1}{2} \ \left( \frac{MASE }{MASE_{Naive_2}} + \frac{sMAPE }{sMAPE_{Naive_2}} \right)
\end{aligned}
\end{equation}


\subsection{Statistical Base Learners}
\label{section::stat_base_models}
This section gives a brief overview of the statistical base learners used in our experimentation.

\begin{description}
    \item[Auto-ARIMA] a standard method for comparing forecast methods' performances. We use the forecasts from an Auto-ARIMA method that uses maximum-likelihood estimation to approximate the parameters \cite{hyndman2008automatic}.
    \item[Theta] the best method of the M3 competition \cite{assimakopoulos2000theta}. Theta is a simple forecasting method that averages the extrapolated Theta-lines, computed from two given Theta-coefficients, applied to the second differences of the time-series.
    \item[Damped  Holt's] exponential smoothing with a trend component modified with a damping parameter $\phi$ imposed on the trend component~\cite{holt2004forecasting,mckenzie2010damped}.
    \item[Comb (or COMB S-H-D)] the arithmetic average of the three exponential smoothing methods: Single, Holt-Winters and Damped exponential smoothing \cite{makridakis2000m3}. Comb was the winning approach for the M2 competition, and was used as a benchmark in the M4 Competition.
\end{description}

\subsection{\gls{ES-RNN} Base Learner}
\label{section::esrnn_base_model}
The \gls{ES-RNN} method uses late fusing of \gls{ES} models with \gls{LSTM} networks to produce more accurate forecasts than either approach. Smyl \cite{ESRNN} describes the three main elements of the \gls{ES-RNN} as deseasonalisation plus adaptive normalisation, generation of forecasts and ensembling.

The first element of Smyl's approach normalises then deseasonalises the series on the fly by \gls{ES} decomposition. \gls{ES} decomposition also computes the level, seasonality, and second seasonality components to integrate them with the \gls{RNN} forecasts. 

Second, Smyl's methodology allocates the predictions of the time-series trends to the \gls{RNN} method since it can model nonlinear relationships. Hybrid forecasting models often exploit the benefits of linear and nonlinear methods by integrating them \cite{zhang2003time, fathi2019time}. 

The final ensembling of the forecasts is from multiple instances of the hybrid model to mitigate parameter uncertainty \cite{petropoulos2018exploring} and take advantage of averaging over model combinations \cite{chan2018some}. In Smyl's methodology, the hybrid instances are trained with the same architecture for independent randomly initialised runs and different subsets of the time-series if computationally feasible. The final forecast for a given series is the average of the forecasts produced by the top-N best models.

\subsubsection{Preprocessing}
In Smyl's methodology, deseasonalisation is achieved on the fly using \gls{ES} models for time-series that are nonseasonal (yearly and daily frequencies), single-seasonal (monthly, quarterly and weekly frequencies) and double-seasonal (hourly frequency) \cite{gardner1985exponential}. In addition, the formulas are updated to remove the linear trend component, which is modelled using the \gls{RNN} instead. The updated formulas are as follows. For the Nonseasonal models:
\begin{equation}
\begin{aligned}
\label{formula:es_nonseasonal}
l_t = \alpha y_t + (1 - \alpha) l_{t-1},
\end{aligned}
\end{equation}
for the Single-seasonality models:
\begin{equation}
\begin{aligned}
\label{formula:es_singleseasonal}
l_t &= \alpha y_t / s_t + (1 - \alpha) l_{t-1},\\
s_{t + K} &= \beta y_t / l_t + (1 - \beta) s_t,
\end{aligned}
\end{equation}
and the Double-seasonality models:
\begin{equation}
\begin{aligned}
\label{formula:es_doubleseasonal}
l_t &= \alpha y_t / (s_t u_t) + (1 - \alpha) l_{t-1},\\
s_{t + K} &= \beta y_t / (l_t u_t) + (1 - \beta) s_t,\\
u_{t + L} &= \gamma y_t / (l_t s_t) + (1 - \gamma) u_t,
\end{aligned}
\end{equation}
where $y_t$ is the value of the series at time step $t$; $l_t$, $s_t$ and $u_t$ are the level, seasonality and second-seasonality components, respectively; $K$ denotes the number of seasonal observations (i.e., four for quarterly, 12 for monthly and 52 for weekly) and $L$ is the number of double-seasonal observations ($168$ for the hourly frequency data.)

Smyl adopts normalisation using the typical approach of constant size, rolling input and output windows to normalise the level and seasonality components produced by the \gls{ES} methods in Eqs. \ref{formula:es_nonseasonal} - \ref{formula:es_doubleseasonal}. The input window size is defined after experimentation, and the output window size is equal to the length of the forecasting horizon. The values of the input and output windows are then divided by the last value of the level of the input window and, if the series is seasonal, additionally divided by the seasonality component. Lastly, the log function is applied to counter the effects of outliers on the learning process. Furthermore, the time-series domain (e.g., micro, macro and finance) are one-hot encoded and presented as the only meta-features to the \gls{RNN} model.

\subsubsection{Forecasts by the \gls{RNN}} 
The \gls{RNN} receives the normalised, deseasononalised and squashed values of level and seasonality together with the meta-features as inputs. The outputs of the \gls{RNN} are then integrated to complete the \gls{ES-RNN} hybridisation in the following way. For the Nonseasonal models:
\begin{equation}
\begin{aligned}
\label{formula:rnn_nonseasonal}
\hat{y}_{t + 1, \dots, t + h} = \exp{(\operatorname{RNN}(\textbf{x}))} \times l_t,
\end{aligned}
\end{equation}
for the Single-seasonal models:
\begin{equation}
\begin{aligned}
\label{formula:rnn_singleseasonal}
\hat{y}_{t + 1, \dots, t + h} = \exp{(\operatorname{RNN}(\textbf{x}))} \times l_t \times s_{t + 1  , \dots  , t + h},
\end{aligned}
\end{equation}
and Double-seasonal models:
\begin{equation}
\begin{aligned}
\label{formula:rnn_doubleseasonal}
\hat{y}_{t + 1, \dots, t + h} = \exp{(\operatorname{RNN}(\textbf{x}))} \times l_t \times s_{t + 1  , \dots  , t + h} \times u_{t + 1 , \dots  , t + h},
\end{aligned}
\end{equation}
where $\operatorname{RNN}(\textbf{x})$ models the linear trend component from the preprocessed input vector $\textbf{x}$, and $h$ is the forecasting horizon. The $l$, $s$ and $u$ components are from the outputs of the \gls{ES} models during the preprocessing step.

\subsubsection{The Architecture}
The \gls{RNN}s of the \gls{ES-RNN} are constructed from dilated \gls{LSTM} \cite{chang2017dilated} stacks, and in some cases, followed by a nonlinear layer and always a final linear layer. Smyl refers to the linear layer as the "adapter" layer since it adapts the size of the last layer to the size of the forecasting horizon, or twice the size of the forecasting horizon for prediction interval (PI) models. The dilated \gls{LSTM}s improve performance using significantly fewer parameters~\cite{chang2017dilated}. Smyl also extends the dilated \gls{LSTM} with an attention mechanism that exposes the hidden states to the weights of the previous states- a horizon equal to the dilation. To achieve this, Smyl embeds a linear two-layer network into the \gls{LSTM}.

\subsubsection{Loss Function and Optimiser} 
The \gls{ES-RNN} implements a pinball loss function to fit the models using Stochastic Gradient Descent. The loss is defined as follows:
\begin{equation}
\begin{split}
\label{formula:pinball_loss}
L_t &= ( y_t - \hat{y}_t )\tau  \text{, \ if} \ y_t \geq \hat{y}_t \\
& = (\hat{y}_t  - y_t  )(1 - \tau)  \text{, \ if} \ \hat{y}_t > y_t,
\end{split}
\end{equation}
where $\tau$ is configured typically between $0.45$ and $0.49$. Smyl notes that the pinball function is asymmetric, and that it penalises values outside a quantile range differently to deal with any biasing, and minimising it produces quantile regression

A level variability penalty is implemented as a regulariser to smooth the level values in the loss function. Smyl notes that this drastically improves the performance of the hybrid method as it can concentrate on modelling the trend instead of over-fitting seasonality-related patterns.
The \gls{ES-RNN}'s level variability penalty is found by: \romannum{1}) compute the log change, i.e., $d_t = \log(y_{t+1}/y_t)$; \romannum{2}) compute the difference of the changes: $e_t = d_{t+1} - d_t$; \romannum{3}) square and average the differences; and \romannum{4}) lastly, the level variability penalty is multiplied by a constant parameter in the range of $50$ - $100$ before adding it to the loss functions.


\subsection{\gls{AVG}}

As a simple benchmark, this study implements a model averaging technique that averages the forecasts of the weak learners \cite{claeskens2008model}. The combined forecast for a set of forecast models $M$, might thus be expressed by the following notation:
\begin{align}
    \hat{y_t} = \frac{1}{n} \cdot \sum_{m=1}^{n} y_{m_t},
\end{align}
where $n$ is the number of models in $M$ and $y_{m_t}$ is the forecast of a model at time step $t$.

\subsection{\gls{FFORMA}}

The \gls{FFORMA} framework adopts a feature-weighted model averaging strategy \cite{montero2020fforma}. A meta-learner learns how effectively a pool of forecasting models are at their task for different regions of the meta-feature space and then combines their predictions based on the learned model weightings. In addition, the \gls{FFORMA} implements the gradient tree boosting model of \gls{XGBoost}, which the authors note is computationally efficient and has shown promising results for other problems \cite{chen2016xgboost}.

Montero-Manso \emph{et al.}'s \cite{montero2020fforma} meta-learning methodology use inputs of nine models and features extracted from the time-series, which measures the characteristics of a time-series: including, but not limited to, features of lag, correlation, the strength of seasonality and spectral entropy.

Montero-Manso \emph{et al.} \cite{montero2020fforma} implement a custom objective function for the \gls{XGBoost} to minimise. \gls{XGBoost} requires both a gradient and hessian of the objective function to fit the model. The functions for their model is derived as follows. The term $p_m(f_n)$ is firstly defined as the output of the meta-learner for model $m$. A softmax transformation is applied to the numeric values to compute the model weights as the probability that each model is the best as:
\begin{equation}
\begin{aligned}
\label{formula:fforma_softmax}
w_m(f_n) = \frac{\exp{(p_m(f_n))}}{\sum_m \exp{(p_m(f_n))}},
\end{aligned}
\end{equation}

where $w_m(f_n)$ is the weight produced by the GB meta-learner for base learner $m \in M$. For each $n$ time-series, the contribution of each method for the \gls{OWA} error is denoted as $L_{nm}$. The weighted average loss function is computed as:
\begin{equation}
\begin{aligned}
\label{formula:fforma_average_loss}
\Bar{L}_n = \sum^{M}_{m=1} w_m(f_n) L_{nm},
\end{aligned}
\end{equation}
where $\Bar{L}_n$ is the weighted average loss over the set of base learners $M$. The gradient of $\Bar{L}_n$ is then computed as follows:
\begin{equation}
\begin{aligned}
\label{formula:fforma_gradient}
G_{nm} = \frac{\partial \Bar{L}_n}{\partial p_m(f_n)} =
w_{nm}(L_{nm} - \Bar{L}_n).
\end{aligned}
\end{equation}
The hessian of the objective function is then finally derived as follows:
\begin{equation}
\begin{aligned}
\label{formula:fforma_hessian}
H_{nm} = \frac{\partial G_n}{\partial p_m(f_n)} \approx w_{nm}(L_{nm}(1 - w_{nm}) - G_{nm}).
\end{aligned}
\end{equation}

In order to minimise the objective function $\Bar{L}$, the functions $G$ and $H$ are passed to \gls{XGBoost}, and the model's hyperparameters are found using Bayesian optimisation on a limited search space that is determined based on preliminary results and rules-of-thumb.

\begin{figure}[htb!]
    \centering
    \includegraphics[width=\columnwidth]{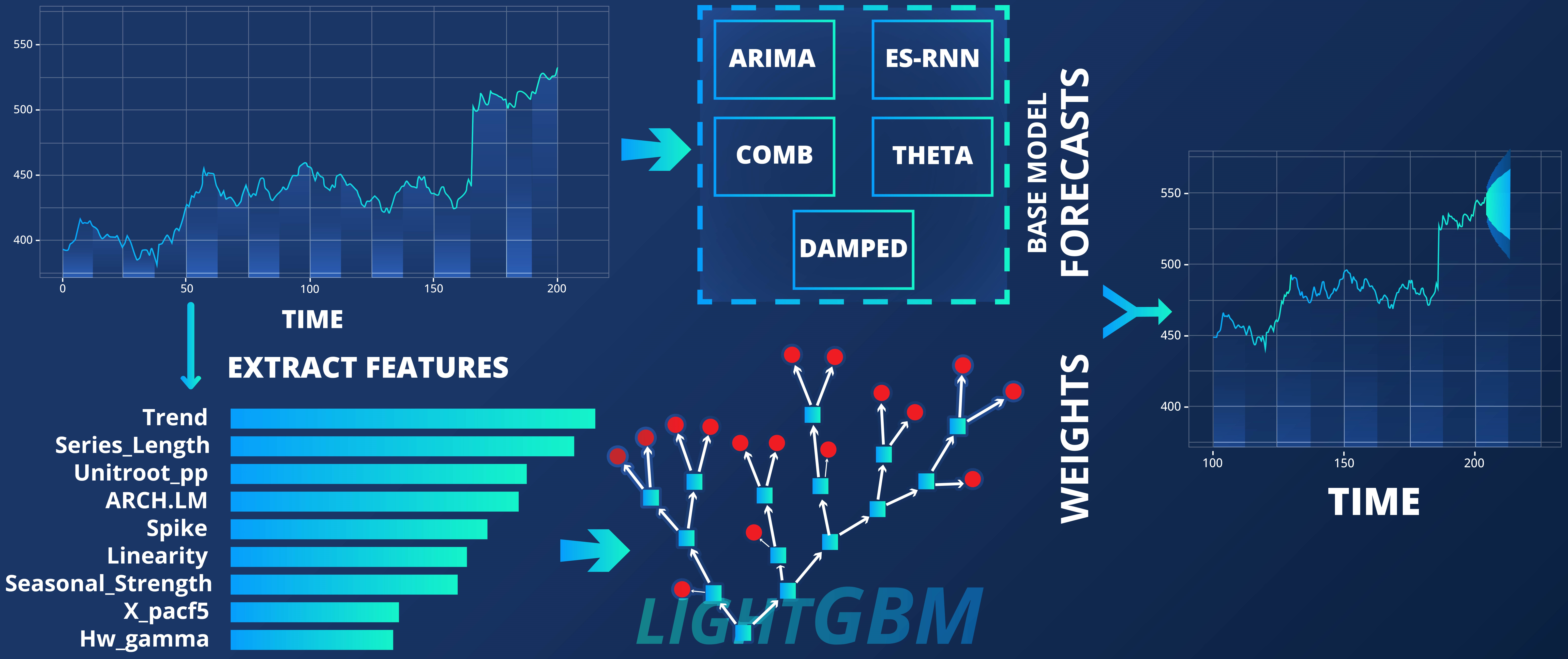}
    \caption{The \gls{FFORMA} forecasting pipeline.}
    \label{fig::framework}
\end{figure}

Montero-Manso \emph{et al.}'s \cite{montero2020fforma} methodology combines the predictions of their pool of forecasting models using Algorithm~\ref{algo:fforma}.
The forecasts are produced from the meta-learner's estimated model weightings, given the meta-features of a new time-series. Thus, the fusion is achieved using the vector of weights $w(f_x) \in \mathbb{R}^M$ with a set of $M$ forecasting models for each time-series $x$. Figure~\ref{fig::framework} depicts this experiment's \gls{FFORMA} forecasting pipeline.

\begin{algorithm}[htb!]
\DontPrintSemicolon 
\SetAlgoLined
\textbf{OFFLINE PHASE: TRAINING} \\
\textbf{Inputs} \\

$\{x_1, x_2, \ldots, x_N\}:$ N observed time-series from the reference set.\\
$\{f_1, f_2, \ldots, f_F\}:$ set of F functions to calculate the meta-features.\\
$\{m_1, m_2, \ldots, m_M\}:$ set of M forecasting models.\\

\textbf{Output}\\
Meta-learner model \\

\textbf{Prepare the metadata} \\
\For{$n \gets 1$ \textbf{to} $N$} {
1. Split $x_n$ into training and test series. \\
2. Calculate  the meta-features $f_n \in F$.\\
3. Fit each base forecasting method $m \in M$ and generate forecasts. \\
4. Compute forecast losses $L_{nm}$ over test period.
}
\textbf{Train the meta-learner} model using the meta-data and base learner forecasting errors, by minimising:

\[ \underset{w}{\argmin{}}\sum^N_{n=1}\sum^M_{m=1} w(f_n)_m L_{nm}.\]
\textbf{ONLINE PHASE: FORECASTING} \\
\textbf{Inputs} \\
The meta-learner model from the offline phase. \\
$\{x_1, x_2, \ldots, x_N\}:$ N observed time-series from validation set.\\
$\{m_1, m_2, \ldots, m_M\}:$ set of M forecasting models.\\

\textbf{Output}\\
$\{y_1, y_2, \ldots, y_N\}:$  Forecast for each series in test set.\\

\For{$n \gets 1$ \textbf{to} $N$} {
1. Calculate meta-features $f_n \in F$. \\
2. Use meta-learner to produce $w(f_n)$, an M-vector of weights. \\
3. Generate forecasts for each $m \in M$. \\
4. Combine the forecasts using $w$. \\
\color{blue}
3-4. FFORMS-G: Select from $M$ the model with the highest allocated $w$ to produce  final forecast. \color{black}
}

\caption{\gls{FFORMA}'s forecast combination \cite{montero2020fforma}.}
\label{algo:fforma}
\color{blue}*FFORMS-G modification of \gls{FFORMA} steps 3 \& 4.
\end{algorithm}
 
The meta-feature for each time-series is computed by a function $f$. The \gls{FFORMA} extracts $43$ meta-features from each time-series using the R package called \emph{tsfeatures} \cite{hyndman2015large}. For nonseasonal time-series, features that only apply to seasonal time-series are set to zero. Unlike the \gls{ES-RNN}, the domain-specific features supplied with the M4 dataset are not utilised in the \gls{FFORMA}.

\subsection{\gls{FFORMS-R}}
\gls{FFORMS-R}, the precursor to \gls{FFORMA}, learns to select a single model from a pool of forecasting models according to their varying performance observed over some meta-data feature space \cite{talagala2018meta}. To this end, \gls{FFORMS-R} uses a random forest ensemble learner \cite{liaw2002classification} to classify a single model as the most relevant for some time-series features extracted from the reference series. Following the original FFORMS paper, we utilise the Gini impurity to determine the quality of a split. The same meta-features are used as per the \gls{FFORMA} approach.

\subsection{\gls{FFORMS-G}}
The \gls{FFORMS-R} framework demonstrated the benefit of selecting a single model from a pool of forecasting models~\cite{talagala2018meta}. We additionally propose replacing the weighted-averaging of the \gls{FFORMA} with the selection of the model with the highest allocated weighting. This replacement allows us to compare how effective model selection against weighted model averaging. This change is highlighted in Algorithm~\ref{algo:fforma}.

\subsection{\gls{NN-STACK}}
Cawood and van Zyl \cite{cawood2021feature} proposed model stacking to do forecasting of nonseasonal time-series. Their implementation performs regression over a feature space consisting of both the ensemble's model forecasts and a set of statistics extracted from the time-series. This paper builds on their tudy by experimenting with more meta-features and a more extensive dataset of different domains and seasonality.

The forecasts of a single timestep is combined using regression over both the ensemble's base learner forecasts and the meta-features extracted from the input series. The \gls{MLP} model performs regression over inputs of the model forecasts and the extracted meta-features, with the target variables as the predicted series' actual values. A neural network of a basic architecture is adopted, and each layer is transformed using the \gls{ReLU} \cite{agarap2018deep} activation function. The model is fit using mini-batches, and the Adam \cite{kingma2014adam} algorithm with a mean absolute error loss function. 

This research extends the original \gls{NN-STACK} implementation by including a more extensive set of meta-features for use by the \gls{FFORMA} framework. Feature selection is achieved using Spearman's rank correlation coefficient ($\rho$) \cite{zar2005spearman}, of the meta-features' correlation with the change in each model performance. This procedure helps reduce problems with overfitting, as only the most correlating features are included in the ensemble.

\subsection{\gls{FFORMA-N}}
We propose an additional \gls{MLP} meta-learner that takes the time-series statistics as inputs and a one-hot encoded vector of the best performing model as the model's targets. The \gls{FFORMA-N} is a deep-learning approach to the \gls{FFORMA}'s feature-weighted model averaging methodology. \gls{FFORMA-N} adopts a softmax activation function in the \gls{MLP}'s output layer to normalise the network's output to a probability distribution, i.e., the probability of each base learner's adequacy to model a time-series. The same meta-features and feature selection procedure are used as per the \gls{NN-STACK} approach.

The neural network is of a deep architecture and adopts a \gls{ReLU} activation function at each layer except for the softmax activated output layer. The model is fit using mini-batches and the Adam stochastic gradient descent algorithm with a categorical cross-entropy loss function. 

The predictions of the forecasting models are combined, similar to the \gls{FFORMA}'s fusion technique. The forecasts from the base learners are summed after they are weighted according to the meta-learner's estimated probability distribution.

\subsection{\gls{N-BEATS}}

Oreshkin \emph{et al.} \cite{oreshkin2019n} reported a 3\% accuracy improvement over the forecasts of the \gls{ES-RNN}. We treat the \gls{N-BEATS} method as a state of the art benchmark to compare the performance of the ensembles. Therefore, the \gls{N-BEATS} method is excluded from the ensembles' pool of base learners. \gls{N-BEATS} is a pure deep learning methodology that does not rely on feature engineering or input-scaling\cite{oreshkin2019n}.

\section{Experimental Method and Results}
\glsresetall
All ensembles are evaluated using ten-fold cross-validation. This validation process is repeated five times, and the scores are averaged over all fifty validation sets to rule out chance from random initialisation. A pseudorandom number generator \cite{blum1986simple} is configured to produce the indices for splitting the data into the training and validation sets for each of the five runs. The seed numbers used were one, two, three, four and five.

All algorithms were implemented in Python and run on a 2.60GHz Intel Core i7 PC with 16 16GB RAM and 1,365 MHz Nvidia RTX 2060 GPU with 6GB GDDR6 memory. The source code of our experiments is available on GitHub \footnote{https://github.com/Pieter-Cawood/FFORMA-ESRNN}.

\subsection{Hyper-parameter Tuning}

\begin{table}[htb!]
\caption{The \gls{FFORMA} hyperparameters}\label{tab::hyper_fforma}
\centering
\begin{tabular}{l|rrrrrr}
Hyperparameter & H & D & W & M & Y & Q \\ 
\bottomrule\toprule
n-estimators      & 2000  & 2000 & 2000 & 1200 & 1200 & 2000 \\ 
min data in leaf  & 63    & 200   & 50  & 100  & 100  & 50  \\ 
number of leaves  & 135   & 94   & 19   & 110  & 110  & 94  \\ 
eta               & 0.61  & 0.90 & 0.46 & 0.20 & 0.10 & 0.75 \\ 
max depth         & 61    & 9   & 17    & 28   & 28   & 43\\ 
subsample         & 0.49  & 0.52 & 0.49 & 0.50 & 0.50 & 0.81 \\ 
colsample bytree  & 0.90  & 0.49 & 0.90 & 0.50 & 0.50 & 0.49 \\ 
\bottomrule
\end{tabular}
\end{table}


Subset names H, D, W, M, Y and Q of Tables~\ref{tab::hyper_fforma} - \ref{tab::owa_results_median} correspond to the Hourly, Daily, Weekly, Monthly, Yearly and Quarterly data subsets. The ensembles' architectures and hyperparameters were found using the training set of the first fold of the first cross-validation run. The two ensembles that use neural networks (\gls{NN-STACK} and \gls{FFORMA-N}) were tuned using backtesting, and the gradient boosting methods (\gls{FFORMA} and \gls{FFORMS-G}) were tuned using Bayesian optimisation. The parameter search space for the Bayesian optimisation was limited based on some initial results, and the Gaussian process method was configured to estimate the parameters that minimise the \gls{OWA} error over $300$ runs of parameter observations.

The \textbf{\gls{FFORMA}} and \textbf{\gls{FFORMS-G}} architecture and hyper-parameters are identical and were automatically tuned using Bayesian optimisation over $300$ runs of a limited parameter search space. For the model's training, early stopping with patience ten was employed using the validation set loss. The validation set is a quarter of the training dataset. Table~\ref{tab::hyper_fforma} presents the hyperparameter settings used to train the \gls{FFORMA} ensembles for the M4 Competition's dataset. 

For the \textbf{\gls{FFORMS-R}}, we used the same hyperparameters to model all subsets of data. The forests were limited to $100$ trees, with each tree having a maximum of sixteen nodes. 

The \textbf{\gls{NN-STACK}'s} architectures and hyperparameters were determined using backtesting on the training dataset of the first fold of the first run of cross-validation. The \gls{MLP} architectures for the Hourly and Weekly subsets were deep with $11$ hidden layers with neuron  in each layer of [$100$, $100$, $100$, $100$, $100$, $50$, $50$, $50$, $50$, $20$ and $20$] sequentially. A more shallow network was used to model the Daily, Monthly, Yearly and Quarterly subsets with three hidden layers with ten neurons each.

Early stopping with patience $15$ and max epochs $300$ was used, except the Daily subset, which had max epochs of $600$. All batch sizes were $225$ except the Daily subset, set at $1200$. Similarly, a learning rate of $0.0003$ was used everywhere except the Daily subset, which used $0.0001$. 

Similar to the stacking method, the \textbf{\gls{FFORMA-N}'s} architecture and hyperparameters were determined using backtesting on the training dataset of the first fold of the first run of cross-validation. The \gls{MLP} architecture was deep, with $11$ hidden layers with number of neurons [$100$, $100$, $100$, $100$, $100$, $50$, $50$, $50$, $50$, $20$ and $20$] sequentially.

Early stopping was configured to terminate the training loop after a configured patience interval of three except for the Daily subset, set at $15$. A maximum number of epochs of $300$ was used except for the Daily subset set to $600$. All batch sizes were $52$ except the Weekly subset at $225$ and the Daily subset, set at $1200$. A learning rate of $0.0001$ was used throughout.


\subsection{Detailed Results}
\begin{figure*}[htb!]
    \begin{subfigure}[B]{.5\textwidth}
    \includegraphics[width=\textwidth]{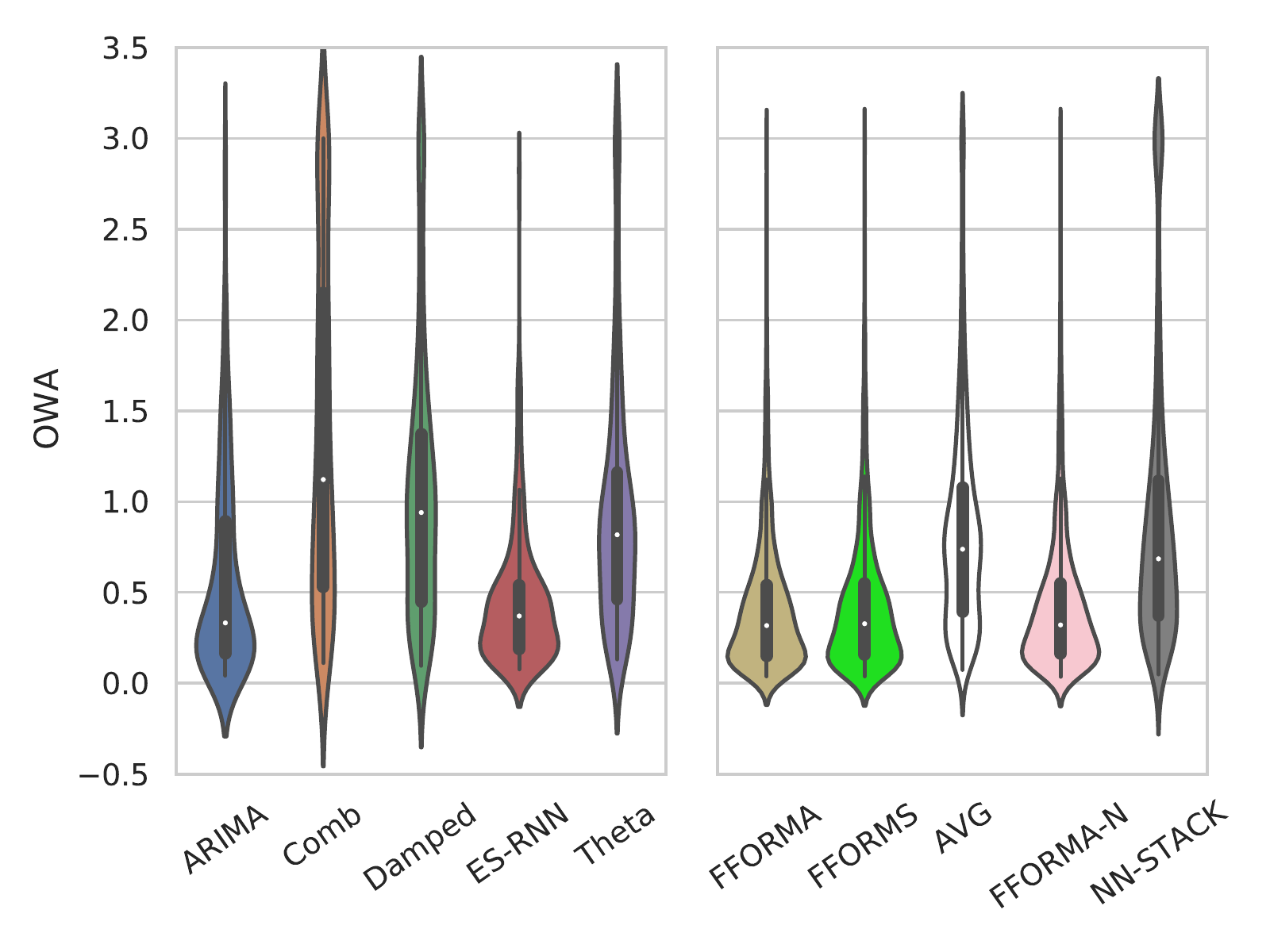}
    \caption{Hourly}\label{fig::violin_hourly}
    \vspace{5pt}
    \end{subfigure}%
    \begin{subfigure}[B]{.5\textwidth}
    \centering
    \includegraphics[width=\textwidth]{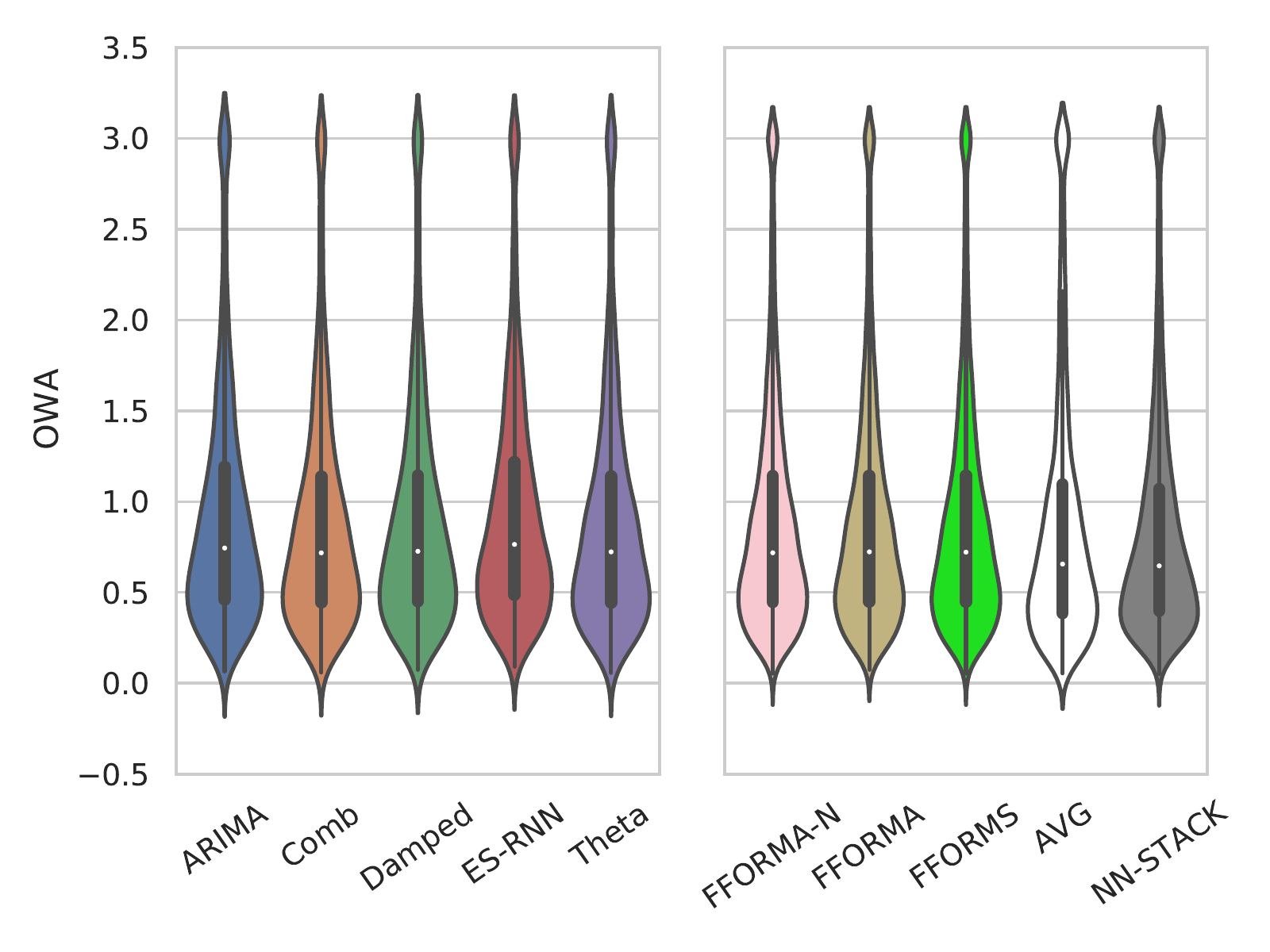}
    \caption{Daily}\label{fig::violin_daily}
    \vspace{5pt}
    \end{subfigure}%
    \\
    \begin{subfigure}[B]{.5\textwidth}
    \includegraphics[width=\textwidth]{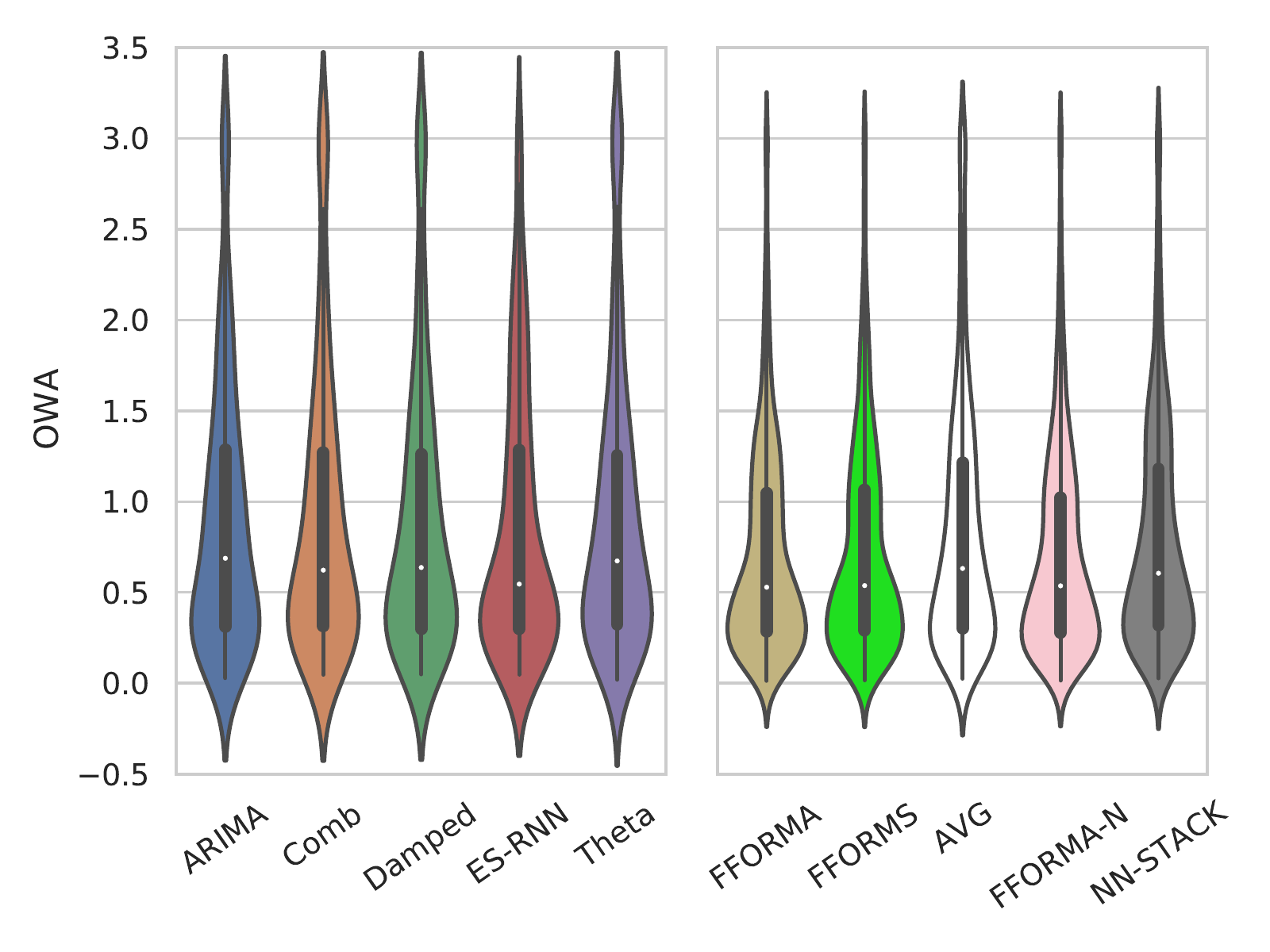}
    \caption{Weekly}\label{fig::violin_weekly}
    \vspace{5pt}
    \end{subfigure}%
    \centering
    \begin{subfigure}[B]{.5\textwidth}
    \centering
    \includegraphics[width=\textwidth]{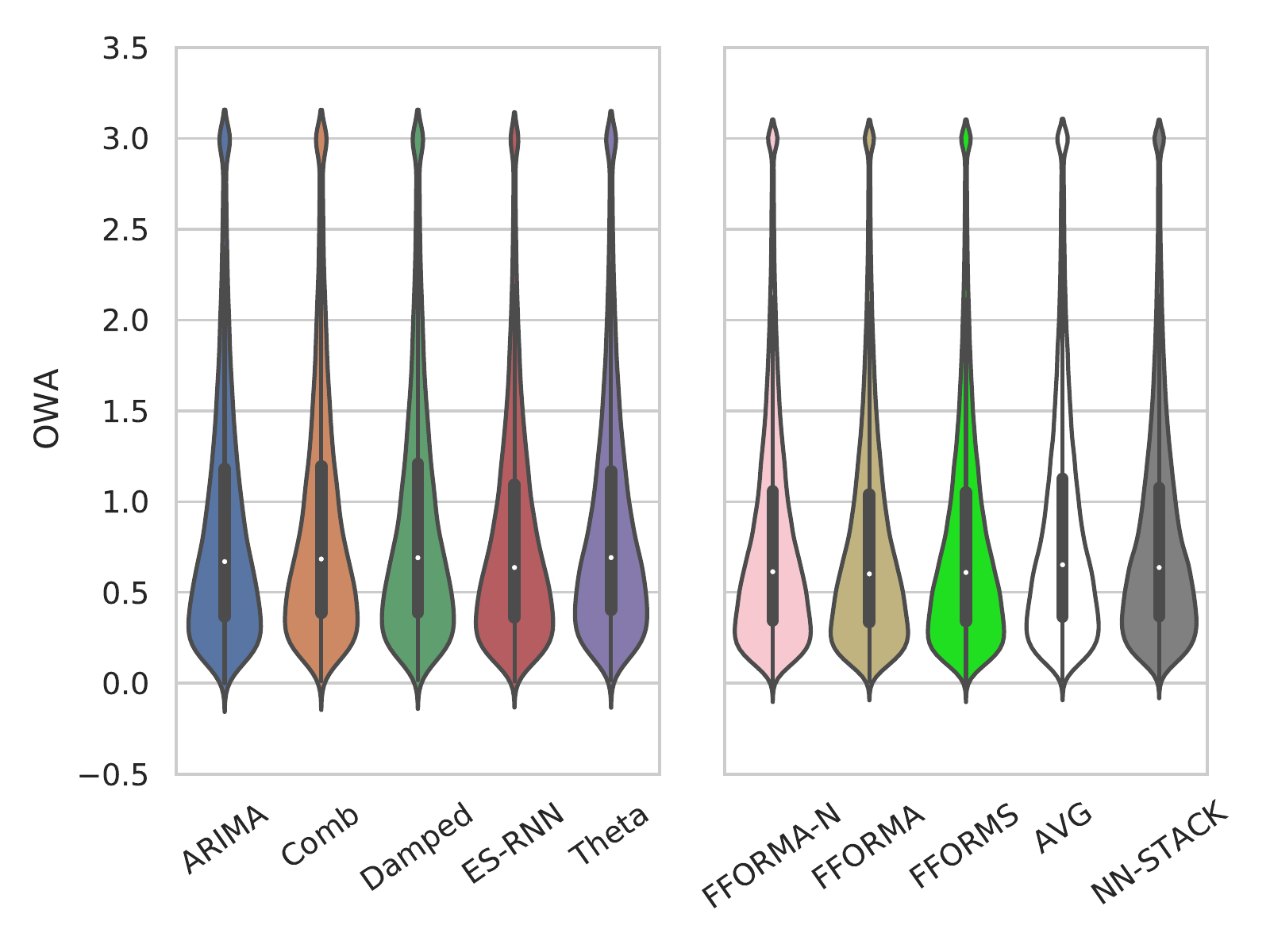}
    \caption{Monthly}\label{fig::violin_monthly}
    \vspace{5pt}
    \end{subfigure}%
    \\
    \begin{subfigure}[B]{.5\textwidth}
    \centering
    \includegraphics[width=\textwidth]{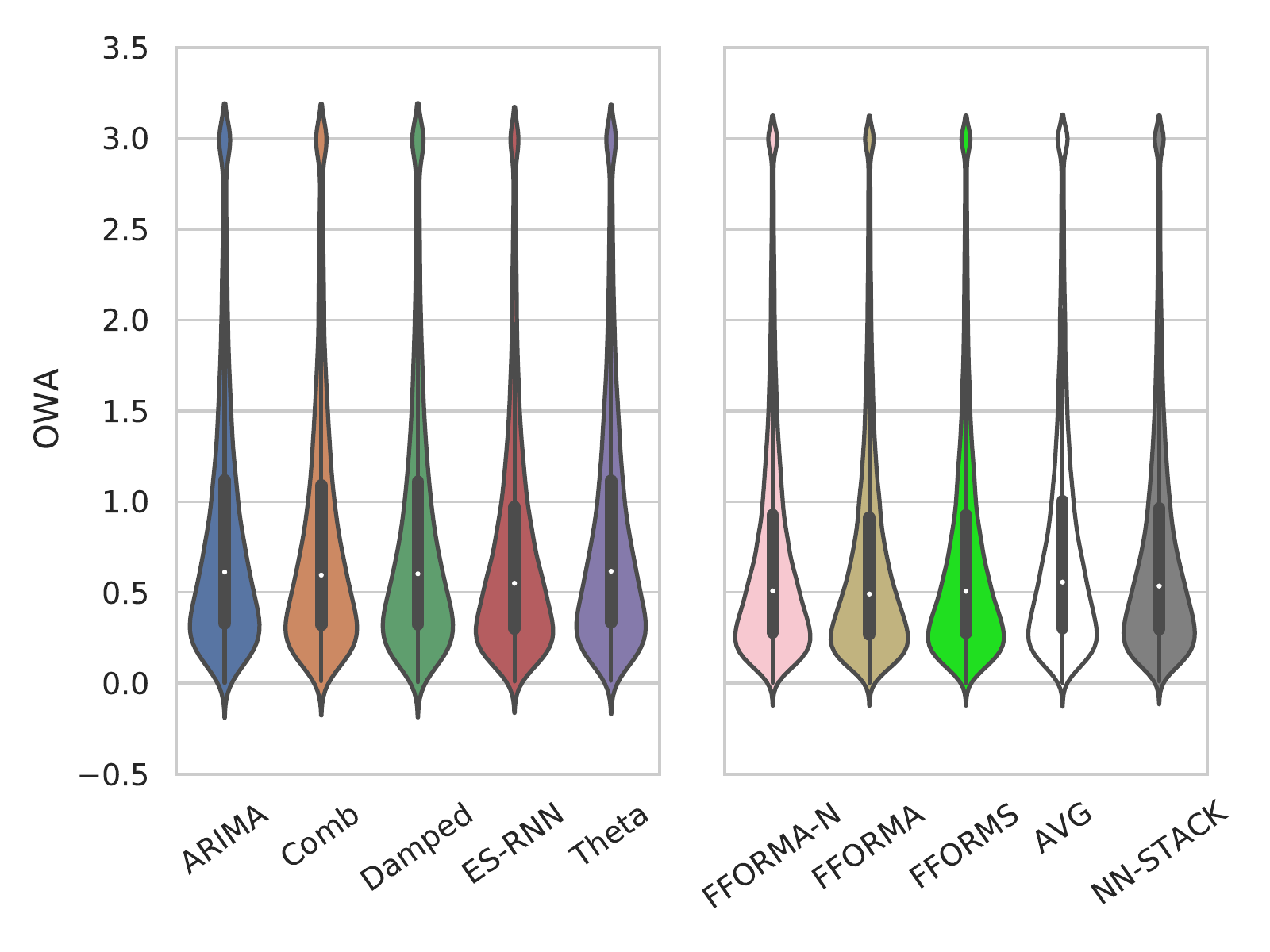}
    \caption{Yearly}\label{fig::violin_yearly}
    \end{subfigure}%
    \begin{subfigure}[B]{.5\textwidth}
    \centering
    \includegraphics[width=\textwidth]{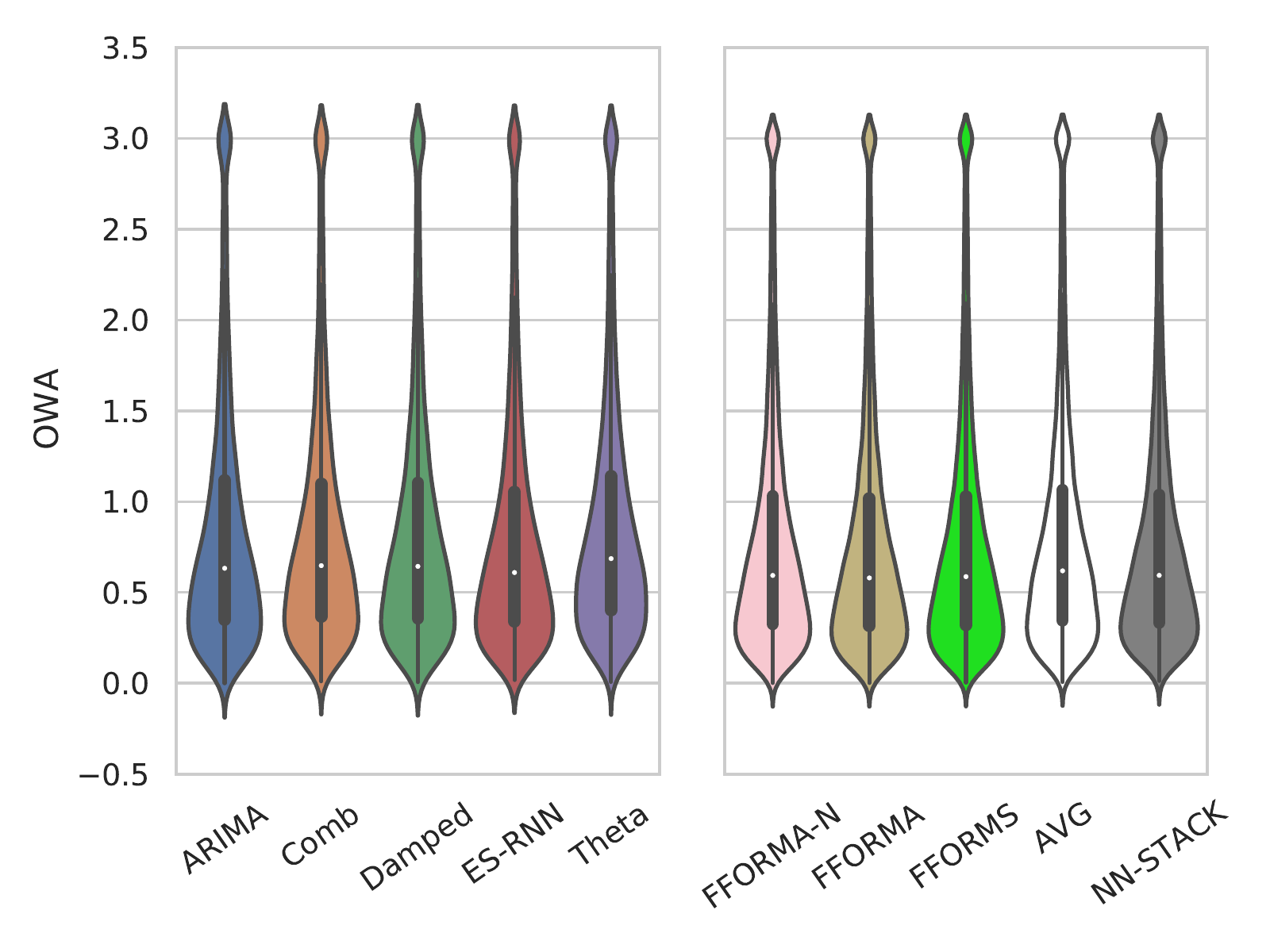}
    \caption{Quarterly}\label{fig::violin_quarterly}
    \end{subfigure}%
    \caption{The \gls{OWA} error distributions on the M4 test set. Where FFORMS represents the similar distributions of \gls{FFORMS-R} and \gls{FFORMS-G}.}
\end{figure*}

\begin{table}[htb!]
\centering
\caption{Average \gls{OWA} errors on the M4 test set. Where lower values are better.} \label{tab::owa_results}
\begin{tabular}{l|rrr|rrrr}
{}&
\makecell[r]{H\\(0.4K)}&
\makecell[r]{D\\(4.2K)}&
\makecell[r]{W\\(0.4K)}&
\makecell[r]{M\\(48K)}&
\makecell[r]{Y\\(23K)}&
\makecell[r]{Q\\(24K)}&
\makecell[r]{M,Y\\\& Q} \\ 
\bottomrule
\multicolumn{8}{c}{\textbf{Base Learners}} \\ 
\toprule
{ARIMA} &{0.577}&{1.047}&{0.925} & {0.903}&{0.892}&{0.898}& 0.899 \\ 
{Comb}  &{1.556}&{0.981}&{0.944} & {0.920}&{0.867}&{0.890}& 0.899 \\ 
{Damped}&{1.141}&{0.999}&{0.992} & {0.924}&{0.890}&{0.893}& 0.907 \\ 
{ES-RNN}&{0.440}&{1.046}&{0.864} & {0.836}&{0.778}&{0.847}& 0.825 \\ 
{Theta} &{1.006}&{0.998}&{0.965} & {0.907}&{0.872}&{0.917}& 0.901 \\ 
\bottomrule
\multicolumn{8}{c}{\textbf{Ensembles}} \\ 
\toprule
{FFORMA}             &\bf{0.415}&{0.983}   &\rf{0.725}&\bf{0.800}&\bf{0.732}&\rf{0.816}& \bf{0.788}\\
{FFORMS-R}           &\rf{0.423}&{0.981}   &{0.740}   &{0.817}   &{0.752}   &{0.830}   & 0.805\\
{FFORMS-G$^\ddagger$}&{0.427}   &{0.984}   &{0.740}   &\rf{0.810}&\rf{0.745}&{0.826}   & \rf{0.798}\\
{AVG}                &{0.847}   &{0.985}   &{0.860}   &{0.863}   &{0.804}   &{0.856}   & 0.847\\
{FFORMA-N$^\ddagger$}&{0.428}   &\rf{0.979}&\bf{0.718}&{0.813}   &{0.746}   &{0.828}   & 0.801\\
{NN-STACK}           &{1.427}   &\bf{0.927}&{0.810}   &{0.833}   &{0.771}   &{0.838}   & 0.819\\ 
\bottomrule
\multicolumn{8}{c}{\textbf{State of the Art}} \\ 
\toprule
{N-BEATS$^\dagger$}&-&-&-&{0.819}&{0.758}&\bf{0.800}& 0.799  \\
\bottomrule
\multicolumn{8}{l}{$\dagger$ reproduced for comparison~\cite{oreshkin2019n}.} \\
\multicolumn{8}{l}{$\ddagger$ proposed methods.}
\end{tabular}
\end{table}

\begin{table}[htb!]
\centering
\caption{Median \gls{OWA} errors on the M4 test set. Where lower values are better.} \label{tab::owa_results_median}
\begin{tabular}{l|rrrrrrr}
{}&
\makecell[r]{H\\(0.4K)}&
\makecell[r]{D\\(4.2K)}&
\makecell[r]{W\\(0.4K)}&
\makecell[r]{M\\(48K)}&
\makecell[r]{Y\\(23K)}&
\makecell[r]{Q\\(24K)}&
\makecell[r]{Schulze\\Rank~\cite{schulze2018schulze}} \\ 
\bottomrule
\multicolumn{8}{c}{\textbf{Base Learners}} \\ 
\toprule
{ARIMA} &{0.332} &{0.745} &{0.688}& {0.670} &{0.612} &{0.633} & 9\\ 
{Comb}  &{1.121} &{0.718} &{0.623}& {0.684} &{0.595} &{0.648} & 8\\ 
{Damped}&{0.940} &{0.727} &{0.637}& {0.691} &{0.602} &{0.644} & 9\\ 
{ES-RNN}&{0.370} &{0.764} &{0.546}& {0.637} &{0.550} &{0.610} & 5\\ 
{Theta} &{0.817} &{0.723} &{0.673}& {0.692} &{0.617} &{0.686} & 11\\ 
\bottomrule
\multicolumn{8}{c}{\textbf{Ensembles}} \\ 
\toprule
{FFORMA}              &\rf{0.318} &{0.723}    &\bf{0.529} &\bf{0.602} &\bf{0.491} &\bf{0.580} & \bf{1}\\
{FFORMS-R}            &\bf{0.311} &{0.711}    &{0.552}    &{0.615}    &{0.511}    &\rf{0.589} & {2}\\
{FFORMS-G$^\ddagger$} &{0.328}    &{0.722}    &\rf{0.538} &\rf{0.610} &\rf{0.506} &{0.596}    & {2}\\
{AVG}                 &{0.738}    &\rf{0.656} &{0.632}    &{0.652}    &{0.557}    &{0.619}    & {7}\\
{FFORMA-N$^\ddagger$} &{0.326}    &{0.720}    &{0.539}    &{0.614}    &{0.508}    &{0.594}    & {2}\\
{NN-STACK}            &{0.685}    &\bf{0.646} &{0.606}    &{0.637}    &{0.535}    &{0.595}    & {5}\\
\bottomrule
\multicolumn{8}{l}{$\ddagger$ proposed methods.}
\end{tabular}
\end{table}

Tables~\ref{tab::owa_results}-\ref{tab::owa_results_median} presents the average and median \gls{OWA} error performance for each forecasting method for each of the six data subsets of the M4 forecasting competition. Column "M, Y \& Q" presents the mean over the three larger subsets. The inclusion of this mean is to provide a comparison to \gls{N-BEATS}. The number of time-series of each data subset is provided in parenthesis, the best results are given in bold, and the top two are highlighted in grey.

Egrioglu and Fildes \cite{egrioglu2020note} showed that the M4 competition's ranking was flawed and that the average \gls{OWA} error measurement is calculated from non-symmetric error distributions and suggest that the errors should be ranked based on their median values instead. Egrioglu and Fildes \cite{egrioglu2020note} also argue that evaluating the combined results is of little value as different methods are ranked best across different subsets. In order to present these more robust results, we present the median \gls{OWA} errors in Table~\ref{tab::owa_results_median}. In addition, Figures~\ref{fig::violin_hourly} - \ref{fig::violin_quarterly} depict the \gls{OWA} error distributions and visualises the data quartiles and extreme values. The visualisations of the error distributions are used to analyse the consistency of each forecast method's accuracy. The violin plots' upper extreme values were clipped to $3.5$, a high \gls{OWA} measurement representing model failure.

\subsubsection{The Hourly Subset}
For the Hourly subset, the \gls{ES-RNN} produced the only stable forecasts amongst the pool of base learners (see Figure~\ref{fig::violin_hourly}). Subsequently, the \gls{NN-STACK} failed as an ensembling method and produced the experiment's worst average ensemble result. This failure was partly due to the small dataset ($414$ series) and the most extensive forecasting horizon requirement of $48$ points. 

The \gls{FFORMA} outperformed all other forecasting methods and produced at least twice as good as the \gls{AVG} benchmark. It was noted here that the \gls{ES-RNN}'s median \gls{OWA} error might slightly ($0.059$) be improved by utilising the hybrid model with the \gls{FFORMA} ensemble technique.

\subsubsection{The Daily Subset} 
The distributions of base learner errors (see Figure~\ref{fig::violin_daily}) are exceptionally similar. Moreover, this was the only subset where the \gls{ES-RNN} failed to outperform other base learners. Subsequently, the gradient boosting methods failed to gain the advantage of the \gls{ES-RNN}'s forecasts as per the other subsets, and the \gls{NN-STACK} method was the most successful ensemble and the best approach.

The results suggest that the \gls{NN-STACK} method is a more suitable ensemble for situations where it is difficult for the ensemble to learn the weightings of superiority amongst the pool of similar performing base learners.

\subsubsection{The Weekly Subset} 
Considering the \gls{OWA} distributions depicted in Figure~\ref{fig::violin_weekly}, the \gls{ES-RNN} produced the lowest \gls{OWA} errors amongst the other base learners, with considerably fewer forecasts reaching the extreme value of $3.0$. This exceptional performance of at least one base learner allowed all ensemble methods to outperform all base learners (see Table~\ref{tab::owa_results}).

The \gls{FFORMA} ensemble improved the \gls{ES-RNN}'s median \gls{OWA} error by a small ($0.017$) margin. The \gls{FFORMA} performed the best of all methods, whereas the \gls{FFORMS-G} was the second-best.

\subsubsection{The Monthly Subset} 
The Monthly subset consisted of $48,000$ time-series and was the largest- and most domain-balanced dataset of all the experiments. Figure~\ref{fig::violin_monthly} depicts the distribution of the \gls{OWA} errors, and the \gls{ES-RNN} performed best amongst the base learners, with a notably smaller number of model failures. These conditions are ideal for analysing the general performance of the ensemble methods, and the results show similar performance in median \gls{OWA} errors and a slight ($0.037$) improvement when the \gls{FFORMA} learns from the predictions of the \gls{ES-RNN}. Nevertheless, considering the median \gls{OWA} error, the \gls{FFORMA} remained the superior method and outperformed all other approaches.

\subsubsection{The Yearly Subset} 
The Yearly subset was one of the experiments' larger and nonseasonal datasets. It was noted that all ensembles, except for the \gls{AVG} benchmark, slightly outperformed the \gls{ES-RNN}. Similar \gls{OWA} distributions were observed between the gradient boosting and neural network ensembles (see Figure~\ref{fig::violin_yearly}.) However, the \gls{FFORMA} showed the greatest ($0.056$) improvement, and it is the preferred ensemble for time-series with a Yearly seasonality.

\subsubsection{The Quarterly Subset} 
The Quarterly subset is similar to the Yearly one in size, balance and forecast horizon. The only exception is that the Quarterly time-series contain a higher seasonality. Considering the ensembles' \gls{OWA} distributions visualised in Figure~\ref{fig::violin_quarterly}, similar ensembling performance was observed to that of the ensembling performance on the Yearly subset. Likewise, no base learner was distinguishable in terms of the \gls{OWA} distributions. However, the more extensive dataset allowed the ensemble methods to avoid the case of the minor Daily dataset (with similar performing base learners), and the \gls{FFORMA} ensemble produced a $0.03$ improvement over the median accuracy of the \gls{ES-RNN}.

The \gls{N-BEATS} approach showed the best average \gls{OWA} result, and it was able to outperform the \gls{FFORMA} for this subset only by a small margin of $0.058$.

\subsection{Overall Results}

\gls{N-BEATS} outperformed all other ensembles only for the Quarterly subset. Therefore, the utility of \gls{N-BEATS} might have notably boosted the ensembles' accuracy for the Quarterly subset. The median \gls{OWA} error gap between the \gls{AVG} and other ensemble methods (e.g., between the \gls{FFORMA}: $0.101$) indicates that Machine Learning is a robust solution to late data fusion.

Although all the ensembles outperform both the \gls{ES-RNN}'s and N-Beats average error when considering the \gls{OWA} error distributions (Figures \ref{fig::violin_monthly} - \ref{fig::violin_quarterly}) of the three larger subsets (Monthly, Yearly and Quarterly), there is still generally a large amount of uncertainty when considering the (Hourly, Daily and Weekly) data sets. Nevertheless, the ensembles were still able to lower the upper quantiles of the \gls{OWA} error distributions by a notable margin. A more significant improvement was observed in the \gls{OWA} error distributions of the three smaller subsets (Hourly, Daily and Weekly) (see Figures \ref{fig::violin_hourly} - \ref{fig::violin_daily}), where the weak learners showed more remarkable performance (considering their \gls{OWA} errors.)

The outstanding ensemble learning performance by both our presented methods and previously by \gls{N-BEATS} provides evidence that:
\begin{enumerate}
    \item the performance of ensemble learning is dependent on the performances of its weak learners;
    \item ensemble learning's performance is dependent on the diversity of the data (i.e., the similarity of the time-series from the different domains.);
    \item for smaller subsets, the traditional methods do better but are still outperformed by the ensemble methods and even more so for larger data sets, where more cross-learning can be exploited;
    \item ensembles of hybrids can still lead to improved performance; and
    \item we reaffirm that there is no free lunch concerning the selection of ensembling methods and that validation will still be required for model selection.
\end{enumerate}

When considering the ensemble learning methods, we note that the gradient boosting ensembles generally outperformed those of neural networks. Furthermore, both gradient boosting ensembles outperformed all other ensemble methods on all subsets except for the Daily subset. Table~\ref{tab::owa_results_median} statistically shows that for all datasets, it is always possible to improve a standalone time-series forecast model by adding it to an ensemble with other statistical forecast models. Lastly, since the \gls{FFORMA} (the second-place M4 submission) utilises the \gls{ES-RNN} as a base learner, the \gls{FFORMA} could outperform the standalone \gls{ES-RNN} (the M4 winner) for all six subsets of data.






\section{Conclusions}
\glsresetall
Our primary objective was to empirically compare ensemble methods against the state of the art forecasting methods, the \gls{ES-RNN} and \gls{N-BEATS}, to determine whether we might \romannum{1}) boost the accuracy of contemporary hybrid models and \romannum{2}) single out a method as the state of the art ensembling technique for future research efforts.

The experiment included four traditional base learners, namely \gls{ARIMA}, Comb, Damped and Theta and an advanced and hybrid machine learning
model, the \gls{ES-RNN}. 
The ensembles of the experiment adopt different ensembling strategies of model averaging, stacking, and selection. We experimented with a random forest meta-learner, namely the \gls{FFORMS-R}, two gradient boosting ones, namely the \gls{FFORMA}, \gls{FFORMS-G}; and two neural network approaches, namely \gls{NN-STACK} and \gls{FFORMA-N}; and a naive arithmetic average as a benchmark. We additionally compare the results of the state of the art benchmark, namely \gls{N-BEATS} using the results reported in their paper.

Regarding the obtained results, we show that the \gls{FFORMA} is a state of the art ensemble technique that might be used to boost the predictive power of powerful methods like the \gls{ES-RNN} and \gls{N-BEATS}. However, considering the case of the Daily subset, the \gls{NN-STACK} approach might be a more suitable ensemble when the pool of forecasting models has similar performance. 
Further, we show that weighted model averaging is a superior ensemble approach as both the \gls{FFORMA} and \gls{FFORMA-N} generally outperform versions of their same architecture that do model stacking or selection.

\gls{NN-STACK} performs regression over single points of the base learner forecasts. \gls{NN-STACK}, therefore, ignores the temporal element of the data, and the meta-learner instead learns a temporally-blind fusion function. Consequently, we recommend that future research consider investigating the performance of a stacking ensemble that is conscious of the data's temporality. Furthermore, considering the equal-weighted averaging adopted by the top submissions of the M5 Competition, further research is needed to assess the performance of feature weighted model averaging for multivariate time-series data.


\vspace{6pt} 



\authorcontributions{Conceptualization, P.C. and T.VZ.; methodology, P.C.; software, P.C.; validation, P.C and T.VZ.; formal analysis, P.C.; investigation, P.C and T.VZ.; resources, P.C.; data curation, P.C.; writing---original draft preparation, P.C.; writing---review and editing, P.C. and T.VZ.; visualization, P.C. and T.VZ; supervision, T.VZ.; project administration, P.C. and T.VZ; All authors have read and agreed to the published version of the manuscript.}

\funding{This research received no external funding.}


\informedconsent{Not applicable.}

\dataavailability{The M4 Competition's dataset and the base learner forecasts can be accessed via the organisers' GitHub repository at \url{https://github.com/Mcompetitions/M4-methods} (accessed on 23 May 2022.)} 


\conflictsofinterest{The authors declare no conflict of interest.} 



\abbreviations{Abbreviations}{
The following abbreviations are used in this manuscript:\\

\noindent 
\printglossaries
 \begin{tabular}{@{}ll}
 ARIMA & Autoregressive integrated moving average\\
 AVG & Model averaging\\
 ES & Exponential smoothing \\
 ES-RNN & Exponential-smoothing-recurrent neural network \\
 FFORMA & Feature-based forecast model averaging \\
 FFORMS & Feature-based forecast model selection \\
 LSTM & Long short-term memory\\
 MASE & Mean absolute scaled error\\
 MLP & Multi layer perceptron\\
 N-BEATS & Neural-basis expansion analysis\\
 FFORMA-N & Neural network model averaging\\
 NN-STACK & Neural network model stacking\\
 OWA & Overall weighted average  \\
 RNN & Recurrent neural network\\
 sMAPE & Symmetric mean absolute percentage error
 \end{tabular}
}




\begin{adjustwidth}{-\extralength}{0cm}

\reftitle{References}

\end{adjustwidth}
\end{document}